\documentclass[12pt]{article}
\usepackage[a4paper,total={18cm,27cm}]{geometry}
\usepackage{graphicx}
\usepackage{comment}
\usepackage{authblk}
\usepackage{amsmath}
\usepackage{amssymb}
\usepackage[utf8]{inputenc}
\usepackage{hyperref}
\usepackage{verbatim}
\usepackage{color,soul}
\usepackage{caption}
\usepackage{xcolor}
\usepackage{framed}
\usepackage{array}
\usepackage{bm}
\colorlet{shadecolor}{yellow!20}

\newcommand{\NI}{\vspace{0.2cm}\noindent}

\newcommand{\is}{\!=\!}
\begin{document}



\title{Illuminating the Black Box
of Reservoir Computing
}


\author[1]{Claus Metzner}
\author[2,3]{Achim Schilling}
\author[2]{Thomas Kinfe}
\author[1]{Andreas Maier}
\author[1,3]{\\ Patrick Krauss}

\affil[1]{\small Cognitive Computational Neuroscience Group, Pattern Recognition Lab, Friedrich-Alexander-University Erlangen-Nürnberg (FAU), Germany}
\affil[2]{\small Neuromodulation and Neuroprosthetics, University Hospital Mannheim, University Heidelberg, Germany}
\affil[3]{\small Neuroscience Lab, University Hospital Erlangen, Germany}

\maketitle


\begin{abstract}
\large
\NI Reservoir computers, based on large recurrent neural networks with fixed random connections, are known to perform a wide range of information processing tasks. However, the nature of data transformations within the reservoir, the interplay of input matrix, reservoir, and readout layer, as well as the effect of varying design parameters remain poorly understood. In this study, we shift the focus from performance maximization to systematic simplification, aiming to identify the minimal computational ingredients required for different model tasks. We examine how many neurons, how much nonlinearity, and which connective structure is necessary and sufficient to perform certain tasks, considering also neurons with non-sigmoidal activation functions and networks with non-random connectivity. Surprisingly, we find non-trivial cases where the readout layer performs the bulk of the computation, with the reservoir merely providing weak nonlinearity and memory. Furthermore, design aspects often considered secondary, such as the structure of the input matrix, the steepness of activation functions, or the precise input/output timing, emerge as critical determinants of system performance in certain tasks. 
\end{abstract}

\newpage
\section{Introduction}

\NI In recent years, deep learning has advanced rapidly \cite{lecun2015deep,alzubaidi2021review}, driven in part by the development of large language models \cite{min2023recent}. These models predominantly rely on feedforward architectures, in which data propagates in a single direction from input to output. By contrast, recurrent neural networks (RNNs) incorporate feedback loops, allowing them to operate as self-sustaining dynamical systems \cite{maheswaranathan2019universality} that maintain internal activity independently of ongoing input.

\NI RNNs are known for several universal capabilities, including their theoretical power to approximate arbitrary mappings \cite{schafer2006recurrent} and emulate general dynamical systems \cite{aguiar2023universal}. Such properties, combined with practical strengths, have sparked ongoing interest in the internal workings of these networks. For instance, RNNs can integrate information across extended time periods \cite{jaeger2001echo,schuecker2018optimal,busing2010connectivity,dambre2012information,wallace2013randomly,gonon2021fading}, and they are capable of forming compact yet expressive internal representations by balancing dimensional compression and expansion \cite{farrell2022gradient}.

\NI Another important direction of research has focused on how RNN dynamics can be shaped and stabilized, particularly under the influence of noise — both internal and external \cite{rajan2010stimulus,jaeger2014controlling,haviv2019understanding,molgedey1992suppressing,ikemoto2018noise,krauss2019recurrence,bonsel2021control,metzner2022dynamics}. RNNs have also gained attention as computational models of brain-like processing \cite{barak2017recurrent}. Especially sparse RNNs with relatively few connections per unit, echoing the architecture of biological neural circuits \cite{song2005highly} have shown advantages in memory retention and representational efficiency \cite{brunel2016cortical,narang2017exploring,gerum2020sparsity,folli2018effect}.

\NI In our previous work, we initiated a systematic study of how structural features influence RNN dynamics, starting from small motifs of three neurons \cite{krauss2019analysis}. Building on this, we demonstrated how large autonomous networks can be shaped through three key parameters: the weight distribution width $w$, the connection density $d$, and the excitation/inhibition balance $b$ \cite{krauss2019weight,metzner2022dynamics}. Additionally, we analyzed how such systems respond to noise, revealing resonance-like effects and stochastic stabilization phenomena \cite{bonsel2021control,schilling2022intrinsic,krauss2016stochastic,krauss2019recurrence,schilling2021stochastic,schilling2023predictive,metzner2024recurrence}.

\NI In our more recent studies, we examined how specific dynamical properties and structural regularities affect the performance of reservoir computing systems. 

\NI First \cite{metzner2025nonlinear}, we systematically varied the degree of neuronal nonlinearity and the strength of recurrent coupling to explore their impact on classification accuracy in synthetic tasks. Surprisingly, even 'calm' reservoirs with minimal internal dynamics and very weak nonlinearity proved capable of generating linearly separable representations for the readout layer, as long as subtle high-dimensional state features were exploited. While task performance generally decreased under strongly chaotic dynamics, accuracy often peaked near the transition zones between chaotic and oscillatory or fix-point regimes, supporting the “edge of chaos” hypothesis.

\NI Second \cite{metzner2025organizational}, we analyzed how biologically inspired structural regularities - such as Dale's principle \footnote{Dale’s principle, originally formulated by Henry Hallett Dale (1935) \cite{Dale1935}, posits that a neuron exerts the same chemical action at all of its synaptic terminals. In modern neuroscience this is usually interpreted as meaning that a neuron releases the same set of neurotransmitters at all of its outputs. Within computational neuroscience, this is often simplified to the assumption that neurons are functionally homogeneous in their postsynaptic effects—that is, a neuron is either purely excitatory or purely inhibitory. While this idealization is helpful for network modeling, it is an oversimplification: some neurons co-release multiple transmitters, and modulatory effects (e.g. GABA being excitatory early in development) complicate the picture. Nevertheless, the excitatory/inhibitory dichotomy derived from Dale’s principle remains a foundational assumption in many theoretical models of neural circuits \cite{strata1999dale}.}, reciprocal symmetry (as in Hopfield networks \cite{hopfield1982neural}), and brain-like modularity \cite{sporns2016modular,meunier2010modular} - shape both the autonomous dynamics of RNNs and their functional performance in reservoir computing. We found that Dale-conform output weights and modular network structure tend to enhance task accuracy, likely by supporting stable and flexible internal representations. In contrast, strong reciprocal symmetry was detrimental, often driving the system into premature saturation and reducing its capacity to differentiate inputs. Together, these findings highlight the potential benefits of incorporating organizational structure into randomly connected networks.

\NI In the present work, we examine in greater detail the interplay between the input matrix, the reservoir, and the readout layer in reservoir computing, with particular attention to parameter settings that are often regarded as secondary. To this end, we introduce a set of artificial prototypical tasks designed primarily for diagnostic purposes, each emphasizing different key aspects of recurrent data processing. For each task, we systematically vary design features and parameter settings to analyze how the components interact and contribute to performance.


\section{Methods}

\subsection{Design of Reservoir Computer (RC)}

\NI The RC consists of an input layer, a recurrent reservoir, and a readout layer. The input data comprises $E$ consecutive episodes, each corresponding, for example, to a pattern to be classified. At each time step $t$, the input layer receives $M$ parallel signals $x_m^{(t)} \in [-1, +1]$. These are linearly transformed by the input matrix $\mathbf{I}$ of size $N \times M$ and injected into the reservoir, as described by Eq.~\ref{yEq}. Each input episode spans $T$ time steps.

\NI The {\bf input layer} consists of the matrix $\mathbf{I}$ and is therefore purely linear. Here we use two variants: In the case of the {\bf dense input matrix}, the elements $I_{mn}$ are drawn independently from a normal distribution with zero mean and standard deviation $w_I$. To study autonomous RNN dynamics, we occasionally set $w_I = 0$, effectively decoupling the reservoir from external input. In the case of the {\bf sparse input matrix}, the $M$ input signals are fed separately to the first $M$ reservoir neurons, using connections of equal strengths $w_I$. Except these first $M$ diagonal entries, the sparse input matrix $\mathbf{I}$ has only zero elements.

\NI A reservoir consists of $N$ recurrently connected neurons, usually with sigmoidal activation functions of the form $y_n=\tanh(u_n)$, where $u_n$ stands for the total weighted and biased input (see argument in Eq.~\ref{yEq}). At each time step, all neuron states $y_n$ are updated in parallel. Each neuron receives a bias term $b_{w,n}$, input from the external signals via $\mathbf{I}$, and recurrent input via a random weight matrix $\mathbf{W}$ (see Eq.~\ref{yEq}). The details of the weight matrix are described below.

\NI For some test cases, we use {\bf non-standard neural activations}, such as scaled sigmoidal functions $y_n=\tanh(s\cdot u_n)$, Gaussian functions $y_n=\exp(-u_n^2)$, sinusoidal functions $y_n=\sin(u_n)$, Heaviside functions $y_n=\Theta(u_n)$, or linear functions $y_n=u_n$. 

\NI Also, the reservoir's random internal connectivity,  described by weight matrix $\mathbf{W}$, is occasionally replaced by {\bf non-standard network structures}: In the case of 'autapse-only' networks, we first create a standard weight matrix and then set to zero all connections between different neurons $n\neq m$, leaving only self-connections $W_{nn}$. In the case of 'neural-loop' networks, each neuron $n$ is only connected to a single successor $n\!+\!1$, and the final neuron $N\!-\!1$ back to $0$.

\NI Before each new episode, the neurons are set to a fixed {\bf initial state} $y_n^{(0)}$, which is either drawn randomly from a uniform distribution in $[-1, +1]$, or simply chosen as zero. This fixed initialization prevents the reservoir from 'contamination' with residual state traces from earlier episodes. 

\NI The {\bf readout layer} performs an affine-linear transformation of the reservoir states $y_n$ using a $K \times N$ output matrix $\mathbf{O}$ and a bias vector $\mathbf{b_o}$, as in Eq.~\ref{zEq}. These parameters are trained via the method of the pseudoinverse (see below). After training, the {\bf continuous readouts $z_k^{(t)}$} can be directly used in tasks with real-valued outputs. For the prediction of binary cellular automaton states $b_k^{(t)}$, we apply a {\bf signum function} to each of the readout channels $k$ (see Eq.~\ref{bEq}). In the case of classification tasks, where a discrete one-hot output $c^{(t)}$ is required, we apply the {\bf argmax function} to the complete readout vector (see Eq.~\ref{cEq}). 

\NI In summary, the {\bf RC is governed by the following equations}:

\begin{eqnarray}
y_n^{(t)} & = & \tanh \left( 
  b_{w,n} + \sum_m I_{nm} x_m^{(t-1)} + 
  \sum_{n^{\prime}} W_{nn^{\prime}} y_{n^{\prime}}^{(t-1)} 
\right) \label{yEq} \\
z_k^{(t)} & = & b_{o,k} + \sum_n O_{kn} y_n^{(t)} \label{zEq} \\
b_k^{(t)} & = &  \operatorname*{sgn} \left( z_k^{(t)} \right)\label{bEq} \\
c^{(t)} & = & \operatorname*{arg\,max} \left\{ z_k^{(t)} \right\}\label{cEq}
\end{eqnarray}


\subsection{Assignment of Reservoir Connection Weights and Biases}

\NI The weight matrix $\mathbf{W}$ of the reservoir's recurrent connections is random but controlled by three statistical parameters: the {\bf density} $d$ of non-zero connections, the excitatory/inhibitory {\bf balance} $b$, and the recurrent {\bf coupling strength} $w$, which is defined by the standard deviation of the Gaussian distribution of weight magnitudes.  

\NI The density $d$ ranges from $d = 0$ (isolated neurons) to $d = 1$ (fully connected network). The balance $b$ ranges from $b = -1$ (purely inhibitory connections) to $b = +1$ (purely excitatory connections), with $b = 0$ corresponding to a perfectly balanced system. The coupling strength $w$ can take any positive value, where $w = 0$ corresponds to unconnected reservoir neurons.  

\NI The bias terms $\mathbf{b_w}$ applied to the reservoir neurons are also drawn from a Gaussian distribution with zero mean and a fixed variance, ensuring that the reservoir operates in a dynamic regime suitable for information processing.

\NI In order to construct an $N \times N$ weight matrix with given parameters $(b, d, w)$, we proceed as follows:  

\begin{enumerate}
    \item First, we generate a matrix $\mathbf{M}^{(magn)}$ of weight magnitudes by drawing the $N^2$ matrix elements independently from a zero-mean normal distribution with standard deviation $w$ and then taking the absolute value.  

    \item Next, we generate a random binary matrix $\mathbf{B}^{(nonz)} \in \{0,1\}^{N \times N}$, where the probability of a matrix element being $1$ is given by the density parameter $d$, that is, $p_1 = d$.  

    \item We then generate another random binary matrix $\mathbf{B}^{(sign)} \in \{-1, +1\}^{N \times N}$, where the probability of a matrix element being $+1$ is given by  
    \begin{equation}
        p_{+1} = \frac{1 + b}{2}
    \end{equation}  
    where $b$ is the balance parameter.  

    \item Finally, the weight matrix is constructed by element-wise multiplication:  
    \begin{equation}
        W_{mn} = M^{(magn)}_{mn} \cdot B^{(nonz)}_{mn} \cdot B^{(sign)}_{mn}
    \end{equation}  
\end{enumerate}

\NI As a standard, we set the density parameter to the maximum value of $d = 1$. The biases $b_{w,n}$ of the reservoir are drawn independently from a zero-mean normal distribution with a standard deviation of $w'$.

\subsection{Standard Reservoir Settings}
\label{sec_standard}
Our so-called {\bf standard reservoir} has a total number of $N=50$ neurons with sigmoidal activation functions of the form $y_n=\tanh(s\cdot u_n)$, using a scaling factor of $s=1$. The neural biases are drawn independently from a Gaussian distribution with zero mean and a standard deviation of $w_B=0.1$. The connection weight matrix has density $d=1$, balance $b=0$, and the weights are drawn independently from a Gaussian distribution with zero mean and standard deviation $w=0.3/\sqrt(N)$. All neurons are reset to value $y_n=0$ at the beginning of each new episode.   


\subsection{Optimal Readout Layer Using Pseudoinverse}

\NI The weights and biases of the readout layer are computed using the pseudoinverse as follows:  

\NI Let \( Y \in \mathbb{R}^{(E-1) \times N} \) be the matrix of reservoir states directly after each input episode, where \( E \) is the total number of episodes and \( N \) is the number of reservoir neurons. Let \( Z \in \mathbb{R}^{(E-1) \times K} \) be the matrix of target output states, where \( K \) is the number of output units.  

\NI To account for biases in the readout layer, a column of ones is appended to \( Y \), resulting in the matrix \( Y_{\text{bias}} \in \mathbb{R}^{(E-1) \times (N+1)} \):  
\[
Y_{\text{bias}} = \begin{bmatrix} Y & \mathbf{1}_{E-1} \end{bmatrix}
\]
where \( \mathbf{1}_{E-1} \in \mathbb{R}^{(E-1) \times 1} \) is a column vector of ones.  

\NI The weights and biases of the readout layer are computed by solving the following equation using the pseudoinverse of \( Y_{\text{bias}} \):  
\begin{equation}
W_{\text{bias}} = Y_{\text{bias}}^+ Z
\label{optWei}
\end{equation}
where \( Y_{\text{bias}}^+ \) is the Moore-Penrose pseudoinverse of \( Y_{\text{bias}} \), and \( W_{\text{bias}} \in \mathbb{R}^{(N+1) \times K} \) contains both the readout weights and the biases.  

\NI To compute the pseudoinverse, we first perform a singular value decomposition (SVD) of \( Y_{\text{bias}} \):  
\[
Y_{\text{bias}} = U S V^\top
\]
where \( U \in \mathbb{R}^{(E-1) \times (E-1)} \) is a unitary matrix, \( S \in \mathbb{R}^{(E-1) \times (N+1)} \) is a diagonal matrix containing the singular values, and \( V^\top \in \mathbb{R}^{(N+1) \times (N+1)} \) is the transpose of a unitary matrix.  

\NI The pseudoinverse of \( Y_{\text{bias}} \) is computed as:  
\[
Y_{\text{bias}}^+ = V S^+ U^\top
\]
where \( S^+ \in \mathbb{R}^{(N+1) \times (E-1)} \) is the pseudoinverse of the diagonal matrix \( S \). The pseudoinverse \( S^+ \) is obtained by taking the reciprocal of all non-zero singular values in \( S \) and leaving zeros unchanged.  

\NI Finally, after inserting \( Y_{\text{bias}}^+ \) into Eq.~\ref{optWei}, the optimal readout weights \( W \in \mathbb{R}^{K \times N} \) and biases \( b \in \mathbb{R}^K \) are extracted from the extended matrix \( W_{\text{bias}} \) as  
\[
W = \left( W_{\text{bias}}^\top \right)_{1:N,\,:}
\]
\[
b_w = \left( W_{\text{bias}}^\top \right)_{N+1,\,:}
\]
where the first \( N \) rows of \( W_{\text{bias}}^\top \) define the readout weights and the last row defines the biases.


\subsection{Fluctuation Measure}

\NI The neural fluctuation measure \( F \) quantifies the average temporal variability of reservoir activations. For each neuron \( n \), we compute the standard deviation \( \sigma_n \) of its activation time series \( y_n^{(t)} \). The global fluctuation is defined as the mean over all neurons:
\[
F = \left\langle \sigma_n \right\rangle_n
\]

\NI Since $\tanh$-neurons produce outputs in \([-1, +1]\), the fluctuation \( F \) lies in \([0, 1]\). A value of \( F = 0 \) indicates a resting or fixpoint state, while \( F = 1 \) corresponds to perfect two-state oscillation (e.g., alternating between $+1$ and $-1$).


\subsection{Correlation Measure}

\NI To assess temporal correlations, we compute the average product of the activation of neuron \( m \) at time \( t \) and neuron \( n \) at time \( t + \Delta t \):
\[
C_{mn}^{(\Delta t)} = \left\langle y_m^{(t)} \cdot y_n^{(t\!+\!\Delta t)} \right\rangle_t
\]

\NI Unlike the Pearson correlation coefficient, we deliberately avoid subtracting the mean or normalizing by the standard deviations. This ensures that the matrix elements \( C_{mn}^{(\Delta t)} \) remain well-defined even when one or both signals are constant, as in a fixpoint state.

\NI The global correlation measure is defined as the average over all neuron pairs:
\[
C_{\Delta t} = \left\langle C_{mn}^{(\Delta t)} \right\rangle_{mn}
\]

\NI Owing to the bounded output of the $\tanh$ neurons, the correlation values \( C_{\Delta t} \) always lie within the range \([-1, +1]\).


\subsection{Nonlinearity Measure}

\NI The shape of the activation distribution \( p(y) \) reflects whether the reservoir operates in a linear or nonlinear regime. A central peak at \( y = 0 \) indicates a linear regime; two peaks near \( \pm1 \) indicate saturation and thus nonlinearity.

\NI We define a nonlinearity measure
\[
\alpha = f_A - f_B + f_C
\]
based on the fractions of neural activations falling into the following intervals:
\[
\begin{array}{ll}
f_A & \in [-1,\;-0.5) \\
f_B & \in [-0.5,\;+0.5] \\
f_C & \in (+0.5,\;+1]
\end{array}
\]

\NI The resulting measure \( \alpha \in [-1, +1] \) distinguishes three regimes:  
$\alpha \approx -1$ for linear operation, $\alpha \approx 0$ for intermediate or flat activation, and $\alpha \approx +1$ for saturated, digital-like behavior.

\NI This intuitive yet robust definition proved most effective among several tested alternatives. It captures the essential qualitative transition in \( p(y) \) from unimodal (linear) to bimodal (nonlinear) distributions, as highlighted in earlier studies.


\subsection{Reservoir Computer as a Sequence-to-Sequence Map}
\label{SeqSeqMap}

\NI In general, we treat the reservoir computer as a trainable mapping from a sequence $X$ of input vectors to a sequence $Z$ of output vectors, where all vector components are real numbers constrained to the fixed range from $-1$ to $+1$:
\[
X \in \left[-1,+1\right]^{TI \times M} \;\longrightarrow\; Z \in \left[-1,+1\right]^{TO \times K}
\]
Here, $TI$ and $TO$ denote the temporal lengths of the input and output sequences, while $M$ and $K$ are their respective vector dimensions, or 'spatial sizes'.

\NI As illustrated in Fig.\ref{fig_Gen}, a new input vector needs one time step to enter the reservoir state, according to Eq.~\ref{yEq}. Each of the vectors $z$ in the output sequence $Z$ is simultaneously computed from the corresponding momentary reservoir state vector $y$, by applying the learned affine-linear transform of the readout matrix and (in some tasks) by further applying the nonlinear signum or argmax function.

\NI It is possible to skip $\Delta T$ time steps before sending the reservoir state to the readout layer, so that the information can be sufficiently processed. However, there will be a tradeoff with the fading memory of the system.


\subsection{Sequence Memorization Task}
\label{SeqMemTas}
\NI The objective of this task is to reproduce the input sequence, similar to an autoencoder, but with a tunable time delay $\Delta T$ (see Fig.\ref{fig_Gen}(a)).

\NI At timestep $t\is0$ of each episode, the reservoir can optionally be reset to a fixed initial state. During the interval from $t\is1$ to $t\is T\!I$, an input vector $\mathbf{x}^{(t)}$ with $M$ components is presented at each time step. These vectors are fed into the reservoir using either a sparse or dense input matrix. The components of $\mathbf{x}^{(t)}$ are independent random numbers, uniformly distributed in the range $\left[-1,+1\right]$.

\NI The output is computed identically across all tasks: Beginning at timestep $t\is\Delta T\!+\!1$, a sequence of $T\!O$ reservoir states, each represented by an $N$-dimensional vector $\mathbf{y}^{(t)}$, is passed to the readout layer and mapped to $K$-dimensional output vectors $\mathbf{z}^{(t)}$.

\NI In this autoencoding task, the readout performs only a learned affine-linear transformation, and we set $T\!O\!=\!T\!I$ as well as $K\!=\!M$.

\NI The resulting set of $T\!I\!\times\!M$ output values $z_i^{(t)}$ is directly compared to the corresponding input values $x_i^{(t)}$. The reconstruction error for component $i$ at time $t$ is defined as $\epsilon_i^{(t)}=z_i^{(t)}-x_i^{(t)}$. From these, the root-mean-square error is calculated as
\begin{equation}
\epsilon_{R\!M\!S}= \sqrt{\frac{1}{T\!I\!\times\!M} \; \sum_i\sum_t \;(\epsilon_i^{(t)})^2}.
\end{equation}

\NI The overall accuracy is then defined by
\begin{equation}
A = \frac{1}{1 + ( \epsilon_{R\!M\!S}/ \Delta z_{\text{tar}})}.
\end{equation}
Here, $\Delta z_{\text{tar}}$ denotes the standard deviation of the target output sequence, which matches that of the input sequence in this task. This accuracy metric yields a typical baseline around $A\approx1/2$, with perfect performance corresponding to $A=1$.


\subsection{Patches Classification Task}
\label{PatClaTas}

\NI In this task, the inputs are two-dimensional vectors $\mathbf{x}$, randomly drawn from the domain $\left[-1, +1\right]^2$. Within this input domain, $N_c$ classes (orange and blue in Fig.~\ref{fig_PCT}(e)) are assigned to a grid of $N_p \times N_p$ square patches in a random but balanced manner, ensuring that each class appears equally often. To enable smooth approximation of decision boundaries, adjacent patches are separated by narrow data-free gaps of width $d_{\text{gap}}$, which contain no samples in either the training or test sets. In our experiments, we use $N_c = 2$, $N_p = 6$, and $d_{\text{gap}} = 0.1$.

\NI For each incoming input vector $\mathbf{x}$, the reservoir state is read out immediately at the same time step in which $\mathbf{x}$ is injected into the reservoir neurons, i.e., with no delay ($\Delta T = 0$). The $K = N_c$ continuous outputs $z_k$ of the trained readout matrix are interpreted as class scores and converted into a one-hot class label via the argmax function.

\NI In this discrete classification task, accuracy $A$ is defined as the fraction of correctly predicted class labels. Given the balanced class distribution, the chance-level accuracy is $A_{\text{chance}} = 1 / N_c$.


\subsection{Cellular Automaton Prediction Task}

\NI In this task, the $M$ components $x_m \!\in\! \{-1, +1\}$ of the input vector represent binary bits, together encoding one of the $2^M$ possible states of an elementary cellular automaton (CA) with $M$ cells and fixed (zero) boundary conditions. The CA evolves according to a specific update rule with Wolfram Rule Number $W\!R\!N$, and the goal of the reservoir computer is to predict the CA's next state from its current state. For this purpose, the continuous outputs $z_k$ of the readout matrix are converted into binary values $b_k \!\in\! \{-1, +1\}$ with the signum function. The number of outputs $K$ is here equal to the number of inputs $M$. In each episode, the CA's initial state is drawn randomly from the full set of $2^M$ possible configurations, with the test set containing states not present in the training set.

\NI To evaluate performance, we count the number $N_{\text{cor}}$ of binary output vectors $\mathbf{b}^{(t)}$ that match the corresponding target vectors exactly in every bit. The accuracy is then defined as $A=N_{\text{cor}}/N_{\text{tot}}$, where $N_{\text{tot}}$ is the total number of predicted CA states. The chance level accuracy is $A\approx 1/(2^M)$ in this task.



\subsection{Sequence Generation Task}

\NI In this task, the reservoir computer receives, at the beginning of each episode, a single $M$-dimensional input vector that resembles one of $N_{DC}$ class-specific prototype vectors $\mathbf{x}_c$, but with small normally distributed fluctuations of standard deviation $\sigma$ added independently to each component. Based on the discrete input class $c$ indicated by the input vector, the goal of the reservoir computer is to generate a predefined, class-specific output sequence consisting of $T\!O$ vectors, each of dimensionality $K$. Prior to the training and test phase, the components of these $N_{DC}$ target sequences are drawn independently and uniformly from the interval $\left[-1,+1\right]$, remaining fixed subsequently.

\NI Once the test data set is processed, we first compute the RMS error $\epsilon_{R\!M\!S}$ between the actual output of the readout matrix and the corresponding target sequences. The accuracy is then defined as
\begin{equation}
A = \frac{1}{1 + \left( \epsilon_{R\!M\!S} / \Delta z_{\text{tar}} \right)},
\end{equation}
where $\Delta z_{\text{tar}}$ is the standard deviation of the target sequences. This accuracy metric yields a typical baseline around $A\approx1/2$, with perfect performance corresponding to $A=1$.


\subsection{PCA-based Visualization of Reservoir States}

\NI Reservoir states $\mathbf{y}^{(t)} \in \mathbb{R}^N$ are projected onto a two-dimensional subspace using Principal Component Analysis (PCA). The reservoir is run over many episodes and all state vectors, excluding the initial reset state, are collected. PCA is applied to this dataset to determine the principal axes of maximal variance. Subsequent state vectors are mapped into this PCA coordinate system, and only the first two principal components are retained for visualization. Points corresponding to different input classes are displayed with distinct colors.


\section{Results}
\subsection{Dynamical Effects of Reservoir Coupling Strength}

\NI In this work, we typically begin each series of numerical experiments with a defined {\bf standard reservoir} (compare Sec.\ref{sec_standard}), from which we systematically vary individual control parameters. This standard reservoir consists of $N\!=\!50$ neurons with $tanh$ activation functions, is fully connected (density $d=1$), and has a balanced number of excitatory and inhibitory weights (balance $b=0$). The weights are drawn from a Gaussian distribution with zero mean and standard deviation $w$, which we refer to as the reservoir coupling strength. Our default value for this crucial parameter is $w=0.3/\sqrt(N)$, a choice that yields good performance across a wide range of tasks (CITE).

\NI The dynamical state of the reservoir is monitored using four measures, each derived from the time series of neural activations $y_n^{(t)}$ (see Methods section for details). The first is the {\bf fluctuation} $F$, a normalized value between 0 and 1 that captures the average amplitude of neural activity. Closely related is the {\bf nonlinearity} $\bm{\alpha}$, which ranges from –1 in the linear regime to +1 in the digital regime. We also compute two unnormalized {\bf correlations}, $\mathbf{C}_0$ and $\mathbf{C}_1$, corresponding to lag times $\Delta t\is0$ and $\Delta t\is1$, respectively. Both range from –1 in oscillatory regimes to +1 in fixed-point regimes. Together, these four simple metrics provide a useful, compact characterization of the reservoir’s dynamical state (CITE).

\NI In earlier work (CITE), we primarily varied the balance parameter $b$ to steer the reservoir dynamics from oscillatory ($b\approx -1$), through chaotic ($b\approx 0$), to fixed-point behavior ($b\approx +1$). In this study, we instead fix the balance at $b\is0$ and focus on tuning the reservoir coupling strength $w$. This approach allows us to revisit and extend previous findings suggesting the existence of a favorable, 'quiescent' regime at low $w$, in which the reservoir remains free of spontaneous fluctuations yet still provides sufficient nonlinearity and mixing for most computational tasks.

\NI To examine the effect of $w$ on the dynamics of the standard reservoir under realistic input conditions, the first 10 neurons receive independent random inputs at time step $t\is1$ of each episode. These input values are uniformly drawn from the interval $\left[-1,+1\right]$. Immediately before this, at $t\is0$, all 50 reservoir neurons are initialized to zero. After the random input vector is applied, the reservoir is allowed to evolve freely for 10 time steps before the next episode begins with a fresh initialization.

\begin{figure}[ht!]
\centering
\includegraphics[width=1\linewidth]{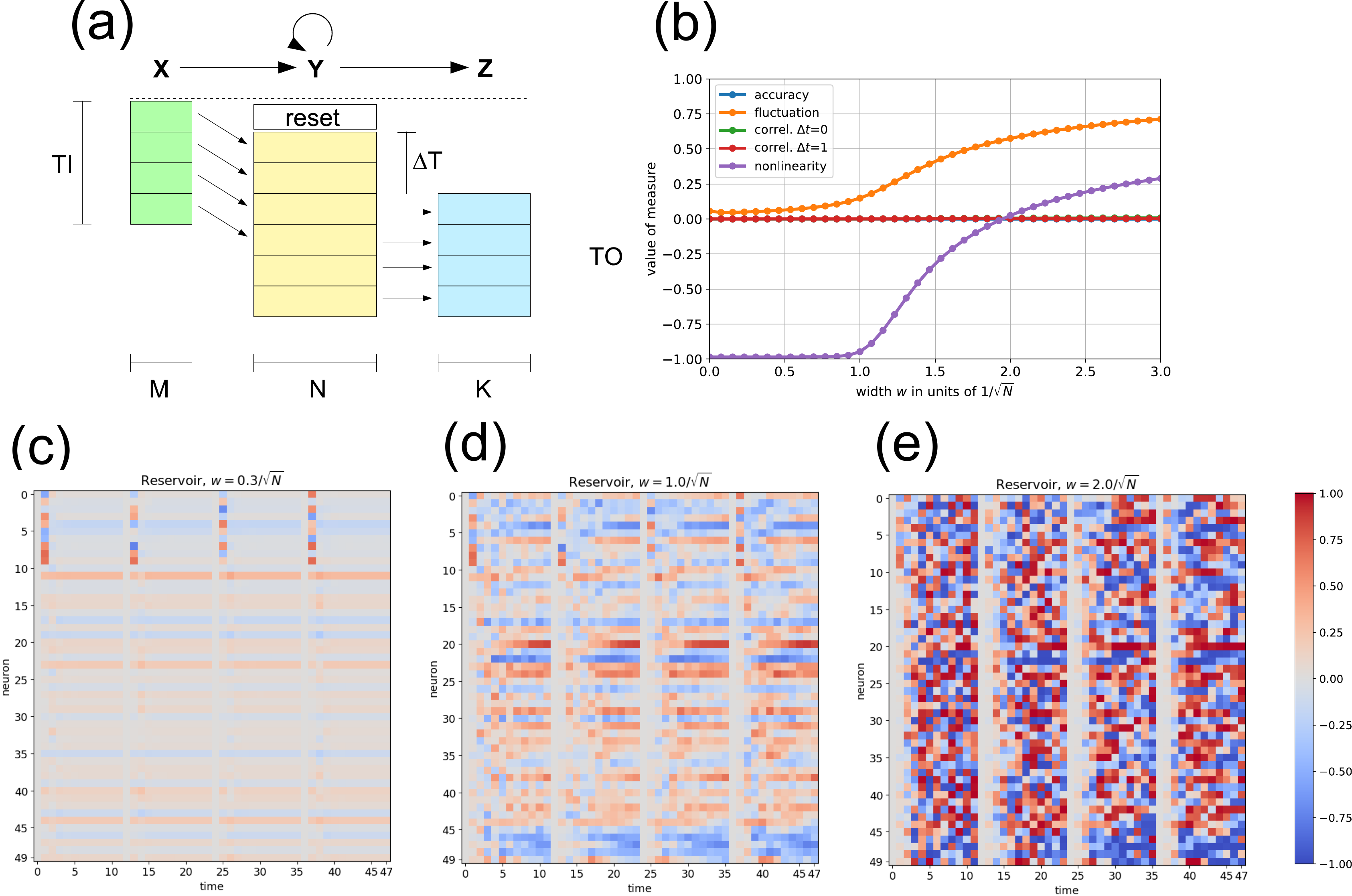}
\caption{
{\bf (a) Information flux in the reservoir computer during one episode}. In the sketch, time steps run from top to bottom, and vector dimensions from left to right. Each colored box represents one vector. The input sequence $\mathbf{X}$ consists of $TI$ vectors, each with $M$ dimensions. Each of these input vectors $x$ (green) becomes a part of the $N$-dimensional reservoir state one time step later. Using the standard settings, the reservoir is reset to the same initial state before the first input of an episode is fed in. After potentially skipping the first $\Delta T$ reservoir states, each subsequent reservoir state vector $y$ (yellow) is send to the readout layer, which instantaneously converts it into an output vector $z$ (blue). The resulting output sequence $\mathbf{Z}$ has a 'temporal height' of $TO$ and a 'spatial width' of $K$. The whole information processing can thus be viewed as a mapping from the input sequence $\mathbf{X}$ to the output sequence $\mathbf{Z}$.
{\bf (b) Dynamical reservoir properties versus neural coupling strength $w$}. We consider a standard reservoir with $N\is50$ neurons, density $d\is1$, and balance $b\is0$. In each episode, at time step $t\is0$, all neurons are set to zero activation. In time step $t\is1$, the first $M\is10$ neurons receive random inputs, and for time steps $t\is2\!\ldots\!11$ the reservoir is updating freely. By averaging over 1000 such episodes, we find that the nonlinearity parameter (magenta) is close to $\alpha=\!-\!1$ for small $w$, indicating almost linear behavior. At around $w\!\approx\!1$, the nonlinearity $\alpha$ and fluctuation $F$ (orange) rise sharply, as spontaneous chaotic fluctuations are driving the reservoir out of the 'quiescent' working regime. The other three dynamical quantities are zero (correlations $C_0$ and $C_1$), or not defined (accuracy) in this simulation. {\bf (c-e) Activations of reservoir neurons versus time}, during a part of the simulations behind panel (b). The activation levels in the range $\left[-1,+1\right]$ are color coded (see color bar at the right side). In each of the three plots, one can see the subsequent reset, input injection, and free dynamical evolution, for a total of 4 complete episodes. {\bf(c)} In the quiescent working regime at small coupling strength $w$, activations return to the network's dynamical fixed point almost immediately after the input injection. {\bf(d)} Close to the critical coupling $w\is1/\sqrt{N}$, relaxation back to the fixed point is slower, but the final reservoir states - immediately before the next reset - are rather similar in each episode. {\bf(e)} In the strong coupling regime, activation amplitudes grow large after each injection, and then show chaotic fluctuations. These state sequences are however fully determined by the input vector at the beginning of each episode.
} 
\label{fig_Gen}
\end{figure}

\NI We begin with the standard reservoir setting $w=0.3/\sqrt(N)\approx0.042$ (Fig.\ref{fig_Gen}(c)) and observe that the reservoir consistently relaxes to the same global fixed-point state toward the end of each episode. This state is characterized by low-amplitude activations that differ between neurons but remain constant over time. Its precise form is determined by the reservoir’s internal weights and biases. Notably, the weakly coupled reservoir converges to this resting state extremely rapidly - typically within just 1–2 time steps after the input vector is applied. The dynamical measures (Fig.\ref{fig_Gen}(b)) confirm that at $w=0.3/\sqrt(N)$, the system operates in a quiescent regime: the fluctuation amplitude $F$ is very small, and $\alpha\approx -1$ indicates strongly linear behavior. The temporal correlations $C_0$ and $C_1$ are both close to zero.

\NI Increasing the coupling strength to $w=1/\sqrt(N)\approx0.141$ (Fig.\ref{fig_Gen}(c)), we observe that the reservoir begins to exhibit sustained fluctuations of larger amplitude following each input injection (see also Fig.\ref{fig_Gen}(b)). Unlike in the quiescent regime, these fluctuations persist throughout the entire episode and do not decay. The final reservoir states at the end of each episode remain similar across trials but are no longer identical. The dynamical measures indicate that $w=1/\sqrt(N)$ marks the onset of a new dynamical regime, characterized by a rapid rise in both fluctuation amplitude and nonlinearity.

\NI Finally, we increase the coupling strength to $w=2/\sqrt(N)\approx0.283$ (Fig.\ref{fig_Gen}(d)). This leads to a pronounced increase in fluctuations, and the nonlinearity measure $\alpha\approx 0$ indicates that the neurons now operate in the saturated, digital regime for a substantial portion of the time. The temporal evolution of the reservoir states appears more chaotic; however, due to the consistent reset at the beginning of each episode, the dynamics remain fully determined by the injected input vector.

\subsection{Sequence Memorization Task}

\NI In a typical reservoir computing system, the readout layer generates its output based solely on the reservoir’s current state. To allow the system to process and integrate temporally sequential data, the reservoir must therefore retain distinct traces of past inputs over an extended period. While it is relatively easy for the readout to reconstruct an input vector presented just one time step earlier, this task becomes increasingly difficult with each additional time step. The recurrent internal dynamics of the reservoir tends to mix and obscure the influence of earlier inputs, progressively degrading the separability of their traces. This challenge is further amplified when new inputs are continuously injected, each one potentially masking the lingering effects of its predecessors.

\NI We test the reservoir’s ability to retain past inputs using a Sequence Memorization Task. In this task, the reservoir computer operates analogously to an autoencoder: a sequence of input vectors is injected into the reservoir, and after a fixed delay $\Delta T$, the system is required to reconstruct the original input sequence with minimal error.

\begin{figure}[ht!]
\centering
\includegraphics[width=0.55\linewidth]{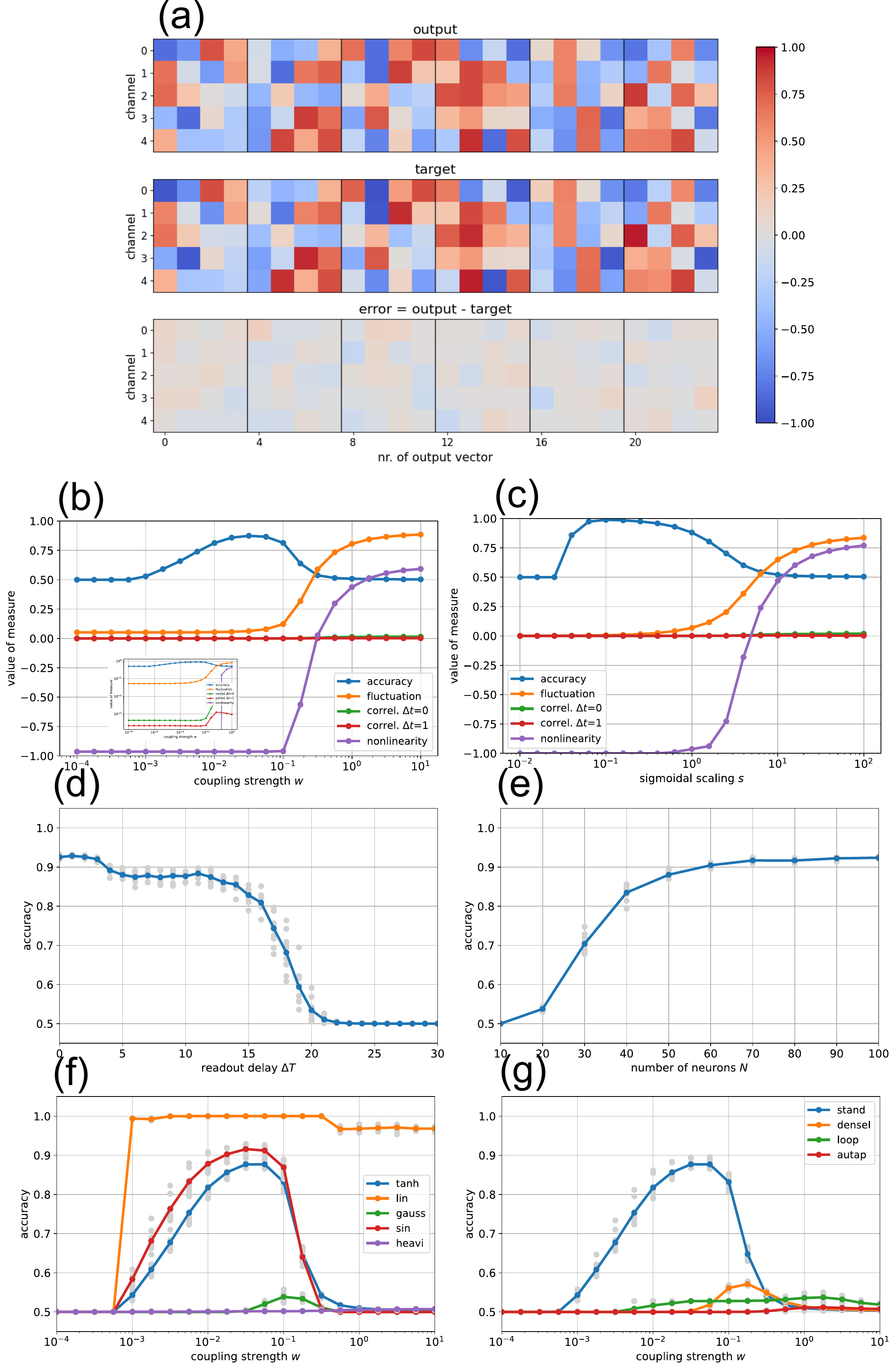}
\caption{
{\bf Sequence Memorization Task} In each episode, the standard reservoir is receiving four input vectors, each consisting of five uniform random numbers in $\left[-1,+1\right]$. After the input sequence is finished, the four vectors are to be reconstructed from the reservoir states. {\bf (a)} Comparing the reservoir computer's actual output (top row) with the target output (middle row), for six subsequent episodes. Time increases from left to right, and the transitions between episodes are marked by vertical lines. The corresponding reconstruction errors (bottom row) are very small, even though the numerical accuracy is only $0.892$ in this example. 
{\bf (b)} Semi-logarithmic plots of the accuracy (blue), together with the dynamical measures of fluctuation (orange) and nonlinearity (magenta), as functions of the reservoir coupling strength $w$. All other parameters are as in the standard reservoir, and the sparse input matrix is used. Plotted quantities are averaged over 10 independent random reservoirs and data sets. The double-logarithmic inset shows that also correlation measures $C_0$ and $C_1$ depend on $w$. 
{\bf (c)} Same quantities as in (b), but versus the scaling parameter $s$ of the neural activation function $y=\tanh(s\cdot u)$. The reservoir computer can perform the sequence memorization task only in a certain range of the control parameters $w$ and $s$. The decline of accuracy for large parameter values is associated with a sharp rise of fluctuation and nonlinearity in the reservoir dynamics.
{\bf (d)} In the standard reservoir, accuracy as a function of the readout delay time $\Delta T$ drops monotonically, but in a step-wise fashion, to baseline level.
{\bf (e)} Accuracy increases monotonically with the number $N$ of neurons in the reservoir, but eventually saturates.
{\bf (f)} Accuracy versus coupling strength $w$ for different neural activation functions (tanh, linear, gaussian, sin and heaviside). Linear neurons work best in this pure memorization task.
{\bf (g)} Accuracy versus coupling strength $w$ for different network topologies of the reservoir computer (standard, dense input matrix, loop network, autapse network). The standard reservoir with sparse input matrix works best.
} 
\label{fig_SMT}
\end{figure}

\NI As an initial test, we inject a sequence of four input vectors, each consisting of five random components, into the standard reservoir of 50 neurons. The input matrix is sparse, targeting only the first five reservoir neurons with the incoming signals. To ensure consistent initial conditions, the reservoir is reset at the beginning of each episode, prior to presenting the first input. The delay $\Delta T$ is set so that reconstruction of the first input vector begins exactly one time step after the final input has been fed into the system (Fig.\ref{fig_SMT}). 

\NI Formally, this autoencoder task can be seen as a mapping from a $4\!\times\!5$-dimensional input sequence to an output sequence of the same shape. However, the mapping is implemented temporally, in a sliding-window fashion: The readout layer receives only the current activations of the 50 reservoir neurons and attempts to reconstruct from this state, instantaneously, the input vector that was presented $\Delta T = 5$ time steps earlier. Although the delay $\Delta T$ is constant across all reconstructed vectors, the first input in each episode may be easier to retrieve, as it follows immediately after a reservoir reset to a well-defined initial state.

\NI We begin with a standard set of 5000 input episodes to train the readout matrix and bias vector using the pseudoinverse method. An independent test set of equal size is then used to evaluate performance by computing the root-mean-square error between the actual and target outputs. From this error, we derive an accuracy metric ranging from 0.5 (chance level) to 1.0 (perfect reconstruction). For details, see the Methods section \ref{SeqMemTas}.

\NI Using the standard reservoir, we achieve an accuracy of $A = 0.892$. While this value may not appear particularly impressive numerically, a visual comparison between the actual and target outputs reveals that the reconstruction performs remarkably well (Fig.\ref{fig_SMT}(a)). When plotted on the same color scale, the reconstruction errors are barely perceptible. Moreover, there is no noticeable difference in error magnitude between the first and last input vector of each episode. This indicates that the affine-linear readout successfully extracts the input vector from $\Delta T$ time steps earlier, regardless of how many random inputs have been injected in the meantime.

\NI We next examine how the accuracy in the sequence memorization task, as well as key dynamical measures, depend on the reservoir coupling strength $w$ (Fig.\ref{fig_SMT}(b)). As before, we use the sparse input matrix. For extremely weak coupling strengths ($w < 10^{-3}$), the accuracy remains at baseline - comparable to that of an untrained readout - presumably because the input signals cannot propagate beyond the five reservoir neurons that receive them directly, and are quickly overwritten by subsequent inputs. As $w$ increases, accuracy improves and reaches a maximum of approximately $A \approx 0.9$ near our standard coupling value of $w \approx 0.042 \approx 0.3/\sqrt{50}$.

\NI As $w$ increases beyond the optimal range, the accuracy declines again. In this same regime of stronger coupling, both the fluctuation and nonlinearity measures rise sharply, indicating that task performance is increasingly impaired by the onset of irregular and chaotic reservoir dynamics.

\NI In the semi-logarithmic plot shown in Fig.\ref{fig_SMT}(b), the two correlation measures $C_0$ (green curve) and $C_1$ (red curve) appear nearly constant, with values close to zero. However, the inset, showing a double-logarithmic plot of the same data, reveals that these quantities increase together with the fluctuation and nonlinearity.

\NI In addition to the coupling strength, another parameter that directly influences the reservoir’s nonlinearity is the scaling factor $s$ in the activation function $y = \tanh(s \cdot u)$. Reducing $s$ below the standard value of $s = 1$ does not change the output range, which remains within $\left[-1, +1\right]$, but it expands the interval of input values $u$ around zero where the activation function behaves approximately linearly.

\NI A semi-logarithmic plot of accuracy $A(s)$ versus the scaling factor $s$ reveals a broad performance peak (Fig.\ref{fig_SMT}(b), blue), similar to the dependence observed for $A(w)$. Remarkably, task accuracy reaches near-perfect levels ($A \approx 1$) at a reduced scaling of $s = 0.1$, at least for this specific task. In contrast, increasing $s$ beyond the standard value of $s = 1$ leads to a decline in accuracy, accompanied by a sharp increase in both fluctuation and nonlinearity measures.

\NI Returning to the standard reservoir settings, we increase the difficulty of the task by extending the readout delay time $\Delta T$ (Fig.\ref{fig_SMT}(c)). As expected, accuracy decreases monotonically with increasing $\Delta T$, approaching baseline around $\Delta T \approx 20$. Still, it is notable that traces of a past input vector can be recovered even after approximately 20 random inputs have been injected in the meantime. Another striking observation is that the accuracy is dropping from $A \approx 0.92$ to $A \approx 0.5$ in a more step-like fashion rather than gradually.

\NI We now reset the delay to $\Delta T = 5$ and investigate how the reservoir size $N$ affects the accuracy $A$, keeping all other parameters at their standard values (Fig.\ref{fig_SMT}(d)). As expected, $A$ increases monotonically with the number of neurons, eventually saturating at around $A \approx 0.92$. For $N \ge 50$, the statistical variability of $A$ across different random reservoir realizations (gray dots) becomes negligible, suggesting that all connection matrices yield similarly good performance once the reservoir is sufficiently large.

\NI In a further experiment, we evaluate the performance of 50-neuron reservoirs using different activation functions (Fig.\ref{fig_SMT}(e)), again plotting the accuracy $A$ as a function of the coupling strength $w$. The standard case using the sigmoidal $\tanh$ function is included as a reference (blue curve). Interestingly, a sinusoidal activation function (red 'sin' curve) outperforms the standard $\tanh$ across a wide range of $w$ values. Most remarkably, the highest accuracy is achieved with purely linear activations (orange 'lin' curve), which yield nearly perfect reconstruction across several orders of magnitude in coupling strength. In contrast, Gaussian activation functions perform poorly (green 'gauss' curve), and only within a narrow window around $w \approx 0.1$. Heaviside activations (magenta 'heavi' curve) fail entirely, as their binary output cannot approximate the continuous target values required by the task.

\NI Finally, we compare the performance of different reservoir network topologies (Fig.\ref{fig_SMT}(e)), again plotting the accuracy $A$ as a function of the coupling strength $w$, with the standard configuration included as a reference (blue 'stand' curve). Replacing the sparse input matrix with a dense one (orange 'dense I' curve) significantly degrades performance in the sequence memorization task. In this case, even the initial input signals immediately activate all neurons, leaving no 'untouched' subpopulation where early input information could propagate once subsequent inputs arrive. A reservoir arranged as a closed linear loop (green 'loop' curve) performs even worse, while networks consisting solely of self-connections (red 'autap' curve) fail entirely.

\subsection{Patches Classification Task}

\NI In general, we view reservoir computers as trainable sequence-to-sequence mappings (see Methods, section~\ref{SeqSeqMap}, and Fig.\ref{fig_Gen}(a)). Within this framework, classification tasks represent a special case in which the output “sequence” consists of a single discrete class label $c$ per episode, while the input sequence may have both spatial width $M$ and temporal depth $T\!I$.

\NI Classification tasks involving temporal structure of the input require the reservoir to retain information over time. Since this memory capacity has already been assessed separately via the Sequence Memorization Task, we now deliberately focus on a non-temporal setting in which each episode consists of a single $M$-dimensional input vector.

\NI When the data contains $K$ distinct classes, the readout layer comprises one linear output unit per class, corresponding to a readout matrix of size $K \!\times\! N$, where $N$ is the number of reservoir neurons. The $K$ continuous outputs $z_k$ of the readout matrix are interpreted as class scores and converted into a single discrete label $c$ via the argmax function.

\NI In this setting, the primary role of the reservoir is to transform the input data into a representation that is linearly separable by the $K$ readout units. In principle, this transformation could require several recurrent updates. However, in simpler cases, the nonlinearity of each neuron, combined with heterogeneous biases and input weights, may already suffice to disentangle the classes without any temporal processing.

\NI Many real-world classification problems involve high-dimensional input vectors, such as the $28\!\times\!28 = 784$-pixel images in the MNIST dataset. However, high input dimensionality does not necessarily imply high classification complexity. For instance, compact Gaussian clusters are easily separable even in very high-dimensional spaces, without requiring nonlinear transformation by the reservoir. Conversely, classification can be difficult even in low-dimensional settings if the class boundaries are highly curved or fragmented.

\begin{figure}[ht!]
\centering
\includegraphics[width=0.8\linewidth]{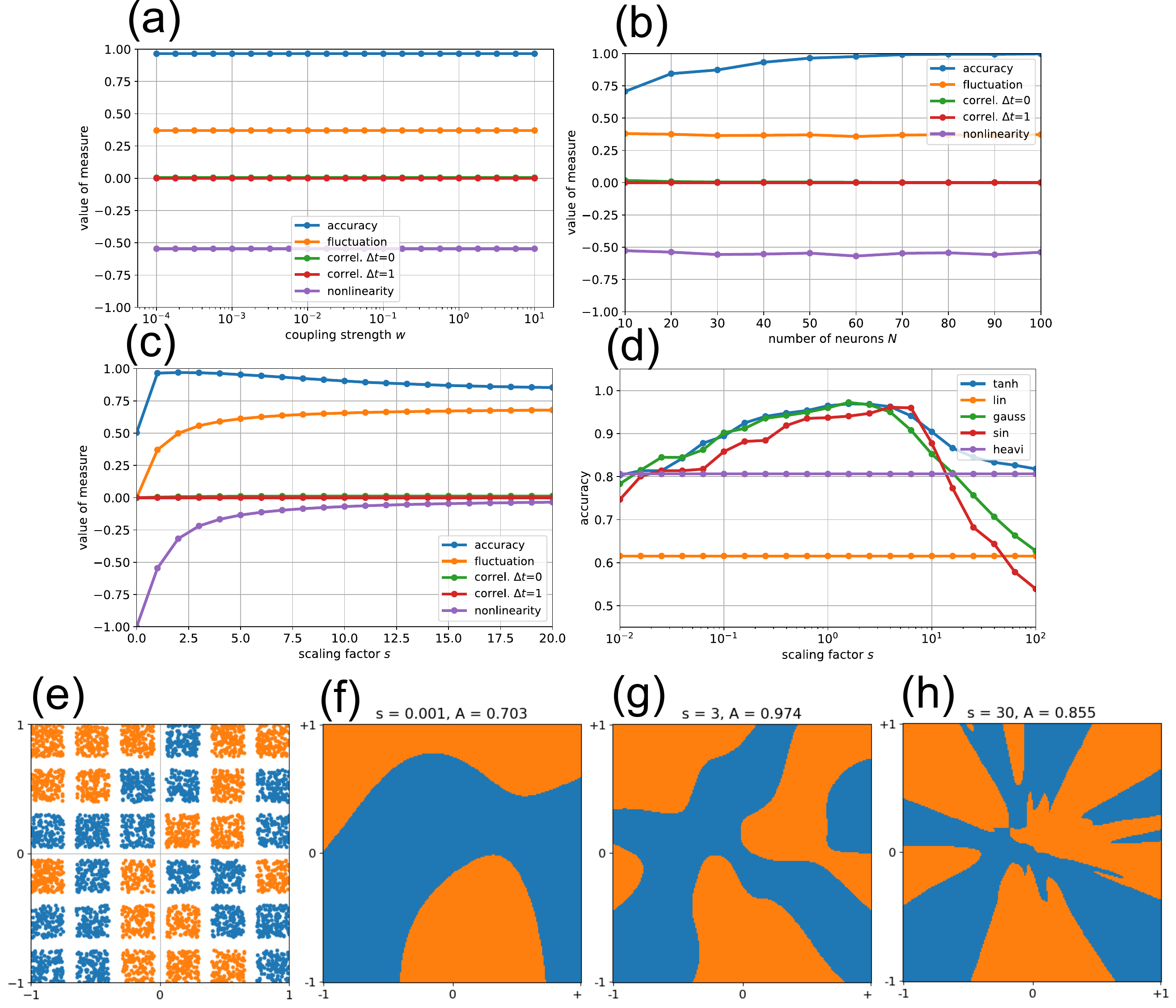}
\caption{
{\bf Patches Classification Task}: As shown in panel (e), two classes (orange and blue) are randomly distributed across a grid of $6\times 6$ square patches within the two-dimensional input area $\mathbf{x}\!\in\!\left[-1, +1\right]^2$. The patches are separated by small gaps of width $d_{gap}=0.1$, which are data-free in the training and test data sets. At the beginning of each episode, two random input signals are injected into all reservoir neurons via a dense input matrix, and the reservoir state is read out immediately to predict the class label (no time delay, $\Delta T = 0$). All results are based on standard reservoir settings, except where parameters are explicitly varied. Plots show averages over 10 random reservoirs and data sets.
{\bf (a)} In this non-temporal classification task, both accuracy and dynamical measures are independent of the coupling strength $w$. The accuracy remains high at $A \approx 0.965$ even when reservoir neurons are uncoupled ($w = 0$), as the readout relies solely on the neurons' nonlinear activation functions.
{\bf (b)} Dynamical measures are also unaffected by the reservoir size $N$. In contrast, accuracy increases monotonically with $N$, as larger reservoirs offer more opportunities for the readout to access neurons with favorable input weights and biases.
{\bf (c)} As a function of the scaling factor $s$, both fluctuation and nonlinearity increase monotonically, reflecting greater sensitivity of neurons to input. Accuracy, however, exhibits a maximum around $s \approx 3$.
{\bf (d)} Semi-logarithmic plot of accuracy versus scaling factor $s$ for different activation functions. Linear (orange) and Heaviside (magenta) activations perform poorly and show no dependence on $s$. For $\tanh$ (blue), Gaussian (green), and sinusoidal (red) activations, accuracy peaks around the same optimal scaling factor, $s_{\text{opt}} \approx 3$.
{\bf (e)} Ground-truth distribution of class labels in the input plane.
{\bf (f–h)} Predicted class label distributions for different values of $s$. At the optimal value $s \approx 3$ (g), the curved decision boundaries closely match the rectangular patch structure of the target. 
} 
\label{fig_PCT}
\end{figure}

\NI To explore this distinction, we focus on the extreme case of two-dimensional input vectors $\mathbf{x} = (x_1, x_2)$, with each component confined to the interval $\left[-1, +1\right]$. Within this input space, we assign two classes in a random but balanced fashion to a grid of square patches (orange and blue in Fig.~\ref{fig_PCT}). Each episode presents only a single input vector, and the reservoir is reset to zero beforehand.
Assuming that multiple reservoir updates are unnecessary in this setting, we read out the reservoir state immediately after the input is injected—i.e., with no delay ($\Delta T = 0$).

\NI Using the standard reservoir configuration but replacing the sparse input matrix with a dense one that feeds both input signals $x_1$ and $x_2$ into all $N = 50$ reservoir neurons, we find that the dynamical properties are entirely independent of the coupling strength $w$ (Fig.~\ref{fig_PCT}(a)). This is expected, as the reservoir is not updated recurrently: in each episode, all neurons start from zero, receive a distinct linear combination of the two inputs (plus individual biases), apply the sigmoidal activation function, and directly produce the reservoir state used for classification and for evaluating the dynamical measures. For the same reason, accuracy is also independent of $w$. In fact, the task achieves a high accuracy of $A \approx 0.965$ even when all recurrent connections are removed ($w = 0$, data not shown).

\NI We also find that the dynamical measures—each defined as an average over neurons or neuron pairs—are independent of the reservoir size $N$, even though in this experiment the coupling strength $w$ was kept at its standard value rather than being scaled  $\propto\!\!1/\sqrt{N}$ (Fig.~\ref{fig_PCT}(b)). In contrast, the average accuracy increases monotonically with $N$, as larger reservoirs provide the readout with more opportunities to exploit neurons whose biases and input connection strengths are particularly favorable. 

\NI Because the Patches Classification Task requires nonlinear bending of decision boundaries in the input plane, we expect that any intervention affecting the degree or form of neuronal nonlinearity will influence the accuracy $A$. Indeed, $A$ varies strongly with the scaling factor $s$, reaching a maximum of nearly $A \approx 1$ at $s \approx 3$ (Fig.~\ref{fig_PCT}(c,d)). Both the fluctuation and nonlinearity measures increase monotonically with $s$, as larger scaling amplifies the total neuronal input and drives the activations further into the saturation regime.

\NI We next compare the performance of different activation functions in this task, again scanning the scaling factor $s$ but now plotting the results on a semi-logarithmic scale (Fig.~\ref{fig_PCT}(d)). As expected, both linear and Heaviside activations fail, as they cannot generate decision boundaries that are both curved and smooth; varying $s$ has no effect on their accuracy. In contrast, the $\tanh$, Gaussian, and sinusoidal activations exhibit similar performance profiles, each reaching peak accuracy within roughly the same range of $s$ values.

\NI Finally, we examine in more detail how the scaling factor $s$ shapes the decision boundaries (Fig.~\ref{fig_PCT}(e–g)). For this analysis, we use a specific dataset (e) in which two classes (orange and blue) are randomly assigned to a $6 !\times! 6$ grid of patches, and compare it with the predicted class-label distribution over the same input domain.

\NI At a very small scaling of $s = 10^{-3}$, where the activation function remains quasi-linear over a broad range around $u = 0$, the reservoir computer can already produce smoothly curved decision boundaries (f). However, both the curvature and the complexity of these boundaries are insufficient to match the target distribution, resulting in a reduced accuracy of $A = 0.703$.

\NI For $s = 3$, corresponding to the peak of $A(s)$, the predicted class distribution shows the best overall agreement with the target (g), achieving an accuracy of $A = 0.974$. The match is not perfect, however, because the rectangular turns in the decision boundaries—needed to follow the square patch layout—cannot be reproduced exactly with smooth, differentiable activation functions.

\NI Increasing the scaling to $s = 30$ produces large derivatives in the activation functions, enabling sharper bends in the decision boundaries. However, this also makes it harder to form gently curved boundaries where required, leading to a drop in accuracy to $A = 0.855$.

\subsection{Cellular Automaton (CA) Prediction Task}

\NI We next consider a prediction task in which the reservoir computer must map a given state $\mathbf{s}^{(t)}$ of a complex deterministic system to its successor state $\mathbf{s}^{(\!t\!+\!1)}$. Candidate systems for such a task include chaotic iterated maps, such as the logistic equation. However, in the chaotic regime, their extreme sensitivity to initial conditions demands exceptionally high numerical precision - requirements for which reservoir computers are generally not well suited.

\NI We therefore employ an elementary cellular automaton (CA) with $M$ cells to generate pairs of initial and successor states. In this setting, each component of the input vector $\mathbf{s}^{(t)}=(s_1^{(t)},\ldots,s_M^{(t)})$ takes a discrete binary value (here implemented as $-1$ or $+1$), thereby avoiding issues of numerical precision. The computation of the successor state $\mathbf{s}^{(\!t\!+\!1)}$, however, still involves nonlinear interactions among all $M$ cell values.

\NI Specifically, we focus on the CA defined by Wolfram’s Rule Number $W\!R\!N\!=\!110$, using fixed boundary conditions. This CA is Turing-complete and computationally irreducible, meaning that distant future states can, in general, only be obtained by iterating through all intermediate steps. In our setting, the test set contains initial states that were never shown during training. Consequently, the only way for the reservoir computer to predict the successor state is to learn the true local update rule, in which each cell determines its next binary state from its own current state and those of its two immediate neighbors, according to a fixed transition table.

\NI We set the CA size to $M\is10$ and begin with the standard reservoir of $N\is50$ neurons, using a dense input matrix. Since the CA has only $2^M = 1024$ possible states, we reduce the training and test set sizes from the standard $5000$ episodes to $1024 / 2 = 512$ each. In every episode, one of the $1024$ possible states is drawn at random and presented as input immediately after the reservoir has been reset to zero; the reservoir state is then passed to the readout layer one update step after input injection. Given the small dataset size, it is virtually certain that the test set contains many states not seen during training (a check shows, for example, $242$ states unique to the training set and $244$ unique to the test set). The resulting accuracy after training is only $A\is0.369$, meaning that perfect prediction of all 10 cells was achieved for just 37 percent of the successor states.

\begin{figure}[ht!]
\centering
\includegraphics[width=0.8\linewidth]{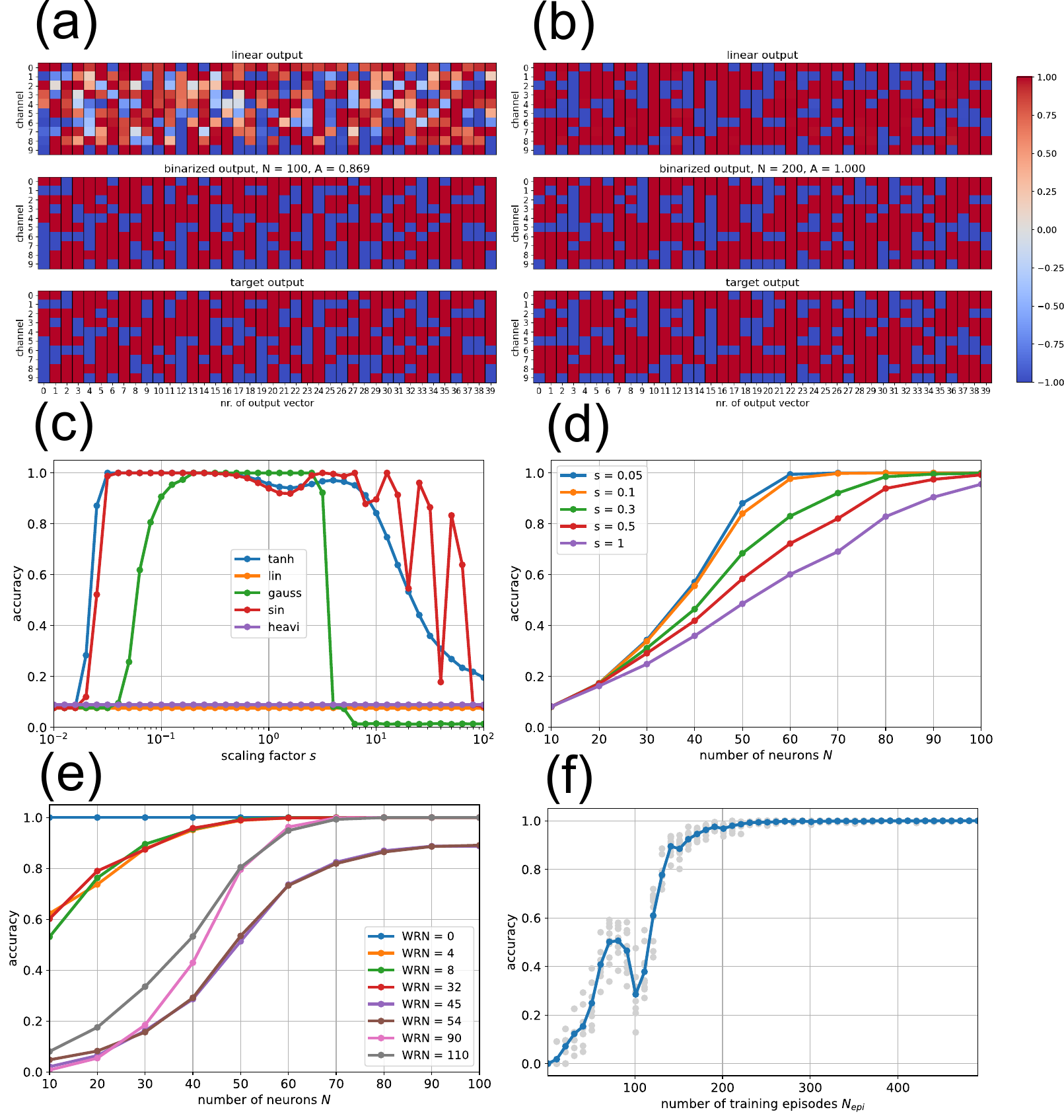}
\caption{
{\bf Cellular Automaton (CA) Prediction Task}: In each episode, the input is a random initial state of Wolfram's elementary CA with $M \!=\! 10$ cells and zero boundary conditions. The task is to predict the subsequent CA state, generated according to rule $W\!R\!N \!=\! 110$.  
{\bf (a)} Comparison of actual and target outputs for 40 episodes using a reservoir with $N \!=\! 100$ neurons and otherwise standard parameters. Each column in the matrix plots represents one predicted or expected successor state. The corresponding initial states are not shown. Top: continuous output of the trained readout matrix. Middle: output after binarization via the sign function, yielding an accuracy of $A \!=\! 0.869$. Bottom: target output.  
{\bf (b)} Same analysis for a reservoir with $N \!=\! 200$ neurons. In this case, the readout matrix produces correct binary outputs directly, achieving perfect accuracy ($A \!=\! 1$) without artificial binarization.
{\bf (c)} Accuracy of a 100-neuron reservoir as a function of the scaling factor $s$ for different activation functions. Linear and Heaviside activations fail entirely, whereas all smooth nonlinear functions can reach $A\is1$ when the scaling $s$ is chosen appropriately.
{\bf (d)} Accuracy as a function of reservoir size $N$ for five different scaling factors. Larger reservoirs generally perform better, but for intermediate sizes around $N\approx50$, reducing $s$ from the standard value of $1$ to $0.05$ leads to a substantial improvement in accuracy.
{\bf (e)} Accuracy as a function of reservoir size $N$ for eight different CA rules, using a scaling factor of $s\is0.1$. Rules 0, 4, 8, and 32, which generate simple dynamics when iterated, are easier to learn than Rules 45, 54, 90, and 110, which exhibit more complex behavior.
{\bf (f)} Accuracy as a function of the number of training episodes $N_{epi}$ for a reservoir with $N\is100$ neurons and scaling parameter $s\is0.1$. Each gray dot represents the result of a single run (specific reservoir and dataset), while the blue line shows the average over ten runs. Overall, accuracy increases with $N_{epi}$, although a noticeable drop occurs around $N_{epi} \approx 100$.
}
\label{fig_CAT}
\end{figure}

\NI Increasing the reservoir size to $N\is100$ (and simultaneously adjusting the coupling strength to $w\is0.3/\sqrt{N}$) raises the accuracy to $A\is0.869$. It is important to note that the readout matrix produces continuous outputs (Fig.~\ref{fig_CAT}(a), top panel). Applying the sign function to these values (middle panel) yields the binary cell states, which are then compared to the target successor states (bottom panel).

\NI Further increasing the reservoir size to $N\is200$ yields perfect performance in the CA prediction task, with $A\is1$. Interestingly, the direct output of the readout matrix is now already almost perfectly binary (Fig.~\ref{fig_CAT}(b), top panel), making the application of the sign function unnecessary in this case.

\NI Returning to a reservoir with $N\is100$ neurons, we next examine the dependence of accuracy $A$ on the scaling factor $s$ for different activation functions (Fig.~\ref{fig_CAT}(c)). As expected, linear activations fail completely. Somewhat surprisingly, Heaviside activations also perform poorly, despite the binary nature of the CA cell states. In contrast, all smooth nonlinear activation functions can achieve perfect accuracy ($A\is1$) even with $N\is100$, provided that $s$ is chosen appropriately. For this specific task, our standard value $s\is1$ is in fact too large for neurons with $\tanh$ activations.

\NI Performance generally improves with increasing reservoir size $N$ (Fig.~\ref{fig_CAT}(d)). However, even the standard reservoir with only $N\is50$ can reach accuracies of about $A\approx0.82$ when the scaling factor is reduced from the standard value $s\is1$ to $s\is0.05$.

\NI For certain combinations of $N$ and $s$, our reservoir computers achieve perfect accuracy ($A\is1$) in predicting the next CA state vector for Rule $110$. In principle, this would allow the output to be fed back as the input for the next episode, enabling computation of CA states arbitrarily far into the future. This corresponds to the emulation of a Turing-complete, deterministic system.

\NI Next, we examine how reservoir computers with a reduced scaling factor $s\is0.1$ and increasing size $N$ perform in predicting the next state of CAs with different Wolfram Rule Numbers (Fig.~\ref{fig_CAT}(e)). Although this task does not involve predicting the long-term evolution of the CAs, we find that rules known to generate simple dynamics, such as $W\!R\!N\in\{0,4,8,32\}$, are easier to learn than rules producing more complex behavior, such as $W\!R\!N\in\{45,54,90,110\}$.

\NI Finally, we examine the accuracy as a function of the number of training episodes $N_{epi}$ for a reservoir with $N\is100$ neurons and scaling parameter $s\is0.1$ (Fig.~\ref{fig_CAT}(f)). Contrary to expectations, the accuracy does not increase monotonically with the size of the training set, but exhibits a pronounced drop at around $N_{epi} \approx 100$. A plausible explanation is overfitting, occurring at a stage where the reservoir computer has still not encountered all distinct training state pairs required to unambiguously infer the underlying local CA update rule.

\subsection{Sequence Generation Task}

\NI We next consider a task in which the reservoir computer must reliably reproduce a predefined target sequence of $T\!O$ output values $z^{(t)}$. The following consideration shows that this should indeed be possible under certain conditions:

\NI The target sequence can be collected into a column vector $\mathbf{z}\in\mathbb{R}^{T\!O}$. At the beginning of each episode, our reservoir is reset to the same initial state, so that the recurrent network thereafter runs deterministically through a sequence of activation vectors $\mathbf{y}^{(t)}$. Stacking these as row vectors yields the state matrix $\mathbf{Y}\in\mathbb{R}^{T\!O\times N}$. A single linear perceptron in the readout layer computes the dot product between each reservoir state $\mathbf{y}^{(t)}$ and a fixed weight vector $\mathbf{o}\in\mathbb{R}^N$, thus generating an output sequence of $T\!O$ values.  
To reproduce the target sequence exactly, the readout weights must satisfy the linear system
$\mathbf{Y}\,\mathbf{o} = \mathbf{z}$.
If the target values are independent random numbers, this is possible in general, provided that the reservoir dimension $N$ is at least as large as the sequence length $T\!O$, and that the subsequent state vectors $\mathbf{y}^{(t)}$ are linearly independent. 

\NI Note that the trajectory $\mathbf{y}^{(t)}$ of the reservoir through state space, and thus also the output sequence $z^{(t)}$, depend on the initial state $z^{(0)}$ used for the reset at the beginning of each episode. If the reservoir does not operate in a chaotic regime, small deviations in $z^{(0)}$ are expected to cause only small errors in the output sequence $z^{(t)}$, rather than being exponentially amplified.

\NI Conversely, if we initialize the reservoir with a completely different state $z^{(0)}$, this will in general produce a different output sequence. In principle, the same reservoir computer could thus be used to generate any one of $N_{DC}$ distinct output sequences, provided it is reset to a corresponding set of $N_{DC}$ distinct initial states. In such a scenario, larger reservoirs are generally required, since a single readout layer has now to accommodate multiple target sequences simultaneously and for this purpose needs sufficiently rich reservoir states.
  
\NI In our practical implementation, the reservoir is always reset to zero neural activations in time step $t\is0$ of each episode. In time step $t\is1$, an $M$-dimensional input vector $\mathbf{x}$ is injected into the first $M$ reservoir neurons, thereby moving the reservoir to a different region of its state space. The subsequent $T\!O$ reservoir state vectors are then passed to the readout to generate the output sequence. Rather than prescribing a simple list of scalar values $z^{(t)}$, the target is defined as a sequence of $T\!O$ output vectors, each of dimension $K$. Accordingly, the readout layer comprises $K$ independent linear perceptrons, each with its own bias term.

\NI We use $K\is2$-dimensional input vectors $\mathbf{x}\in\mathbb{R}^2$. Prior to training, $N_{DC}\is3$ cluster centers $\mathbf{x}_c$ are chosen randomly within $\left[-1,+1\right]^2$. Actual input vectors $\mathbf{x}$ are generated by first selecting a class label $c$ and then adding to the components of $\mathbf{x}_c$ normally distributed random numbers drawn from $\mathcal{N}(0,\sigma^2)$. The resulting vectors are injected into a standard or modified reservoir of $N\is50$ neurons via a sparse input matrix. Readout begins immediately upon input injection, producing a sequence of $T\!O\is10$ output vectors, each of dimension $K\is5$. Performance is evaluated by computing the RMS error between actual and target outputs and converting it into an accuracy measure that ranges from $A\approx0.5$ (untrained) to $A\is1$ (perfect).

\begin{figure}[ht!]
\centering
\includegraphics[width=0.9\linewidth]{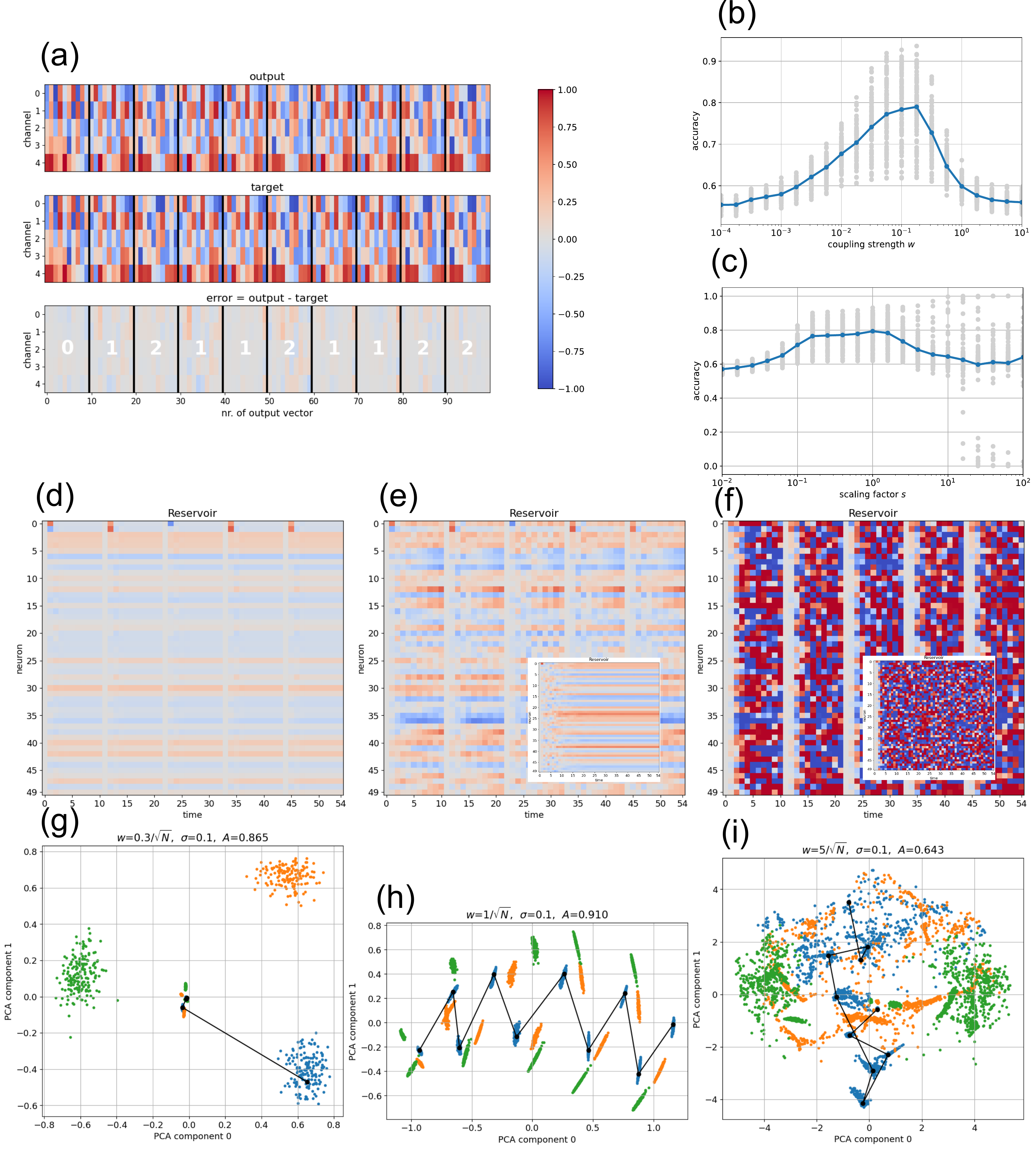}
\caption{
{\bf Sequence Generation Task}: After injection of a 2D input vector $\mathbf{x}$, drawn from one of three Gaussian clusters (classes), the reservoir computer has to generate a class-dependent, predefined sequence of 10 output vectors, each 5-dimensional.  
{$\,\!$\bf (a)} Top: Produced output sequences during 10 episodes, separated by black vertical lines. Middle: Corresponding target sequences. Bottom: Output error. White numbers indicate class labels.
{$\,\!$\bf (b)} Accuracy versus coupling strength $w$ has a clear peak at around $w\approx0.2$, yet statistical fluctuations among data sets are large.
{$\,\!$\bf (c)} Accuracy versus scaling factor $s$ has a soft peak at around $s\approx1$. 
{$\,\!$\bf (d,e,f)} Reservoir states during five episodes for standard weak coupling $w\is0.3/\sqrt{N}$ (d), for critical coupling $w\is1/\sqrt{N}$ (e), and for strong coupling $w\is5/\sqrt{N}$ (f). The insets demonstrate that the long-time dynamical attractor is still a fixed point for critical coupling (e), but becomes chaotic for strong coupling (f). 
{$\,\!$\bf (g)} For $w\is0.3/\sqrt{N}$, the accuracy is $A\is0.865$. Shown is the distribution of reservoir states over 500 episodes in the plane of the first two PCA components, colored by input class. The black trajectory (one episode) relaxes exponentially from the initial state inside the blue cluster toward the reservoir’s final resting state.  
{$\,\!$\bf (h)} For $w\is1/\sqrt{N}$, the accuracy increases to $A\is0.910$. The black trajectory now shows oscillatory behavior, while the three classes remain well separated.  
{$\,\!$\bf (i)} For $w\is5/\sqrt{N}$, the accuracy decreases to $A\is0.643$. The sample trajectory is irregular and classes overlap, indicating a more chaotic dynamical regime.  
}

\label{fig_SGT}
\end{figure}

\NI For the standard coupling strength $w\is0.3/\sqrt{N}$ and a cluster width of $\sigma\is0.1$, we obtain an accuracy of $A\is0.865$ in the specific dataset shown. Even with this modest score, the errors remain very small on the $\left[-1,+1\right]$ scale of the target values (Fig.~\ref{fig_SGT}(a)).

\NI The accuracy as a function of the coupling strength $w$ shows a peak at $w\approx0.2$. The fluctuations of the accuracy among datasets are however large, as the three clusters can be well separated or close together (Fig.~\ref{fig_SGT}(b)). As a function of the scaling factor we find a weak maximum of the accuracy at $s\approx1$ (Fig.~\ref{fig_SGT}(c)).

After each input injection, the neural activations relax rapidly toward the reservoir’s resting state (Fig.~\ref{fig_SGT}(d)). A two-dimensional PCA visualization shows that the first update step already carries the reservoir almost entirely from the post-injection state toward the global resting attractor, while subsequent steps are much shorter, indicating fast exponential convergence (Fig.~\ref{fig_SGT}(g)).

\NI Increasing the coupling strength to $w\is1/\sqrt{N}$ produces stronger activations and more complex dynamics (Fig.~\ref{fig_SGT}(e)). The reservoir still converges to a global fixed point attractor, provided the transient is not interrupted by a new episode (Inset of Fig.~\ref{fig_SGT}(e)). PCA analysis shows that trajectories after input injection no longer approach the attractor exponentially, but in an oscillatory manner (Fig.~\ref{fig_SGT}(h)). Nevertheless, trajectories starting from different clusters remain well separated. In this medium coupling regime, accuracy increases to $A\is0.91$, exceeding that of the weak coupling case.

\NI A further increase of the coupling strength to $w\is5/\sqrt{N}$ pushes the neurons into saturation (Fig.~\ref{fig_SGT}(f)). Activation patterns then differ markedly between episodes, even for identical input clusters. The reservoir exhibits irregular dynamics that persist even after many uninterrupted updates (Inset of Fig.~\ref{fig_SGT}(f)), indicating chaotic behavior. PCA analysis shows that state distributions from different input clusters overlap, at least in the plane of the first two components (Fig.~\ref{fig_SGT}(i)). The readout cannot exploit such irregular trajectories, and accuracy drops to $A\is0.643$.

\section{Discussion}

\NI In this work, we systematically examined the interplay of input matrix, reservoir, and readout layer in reservoir computing, focusing on simplification rather than performance maximization.

\NI We found that reservoirs with very weak internal coupling—the “quiescent” regime—can still support accurate computation by providing fading memory and weak nonlinearity, without requiring strong spontaneous dynamics.

\NI Design aspects often considered secondary—such as the structure of the input matrix, the steepness of activation functions, or the choice of non-standard nonlinearities—proved to be decisive determinants of performance in several tasks.

\NI In the following, we summarize the impact of these design features across the different tasks.

\subsubsection*{Coupling Strength $w$}

\NI The coupling strength $w$ between reservoir neurons is a crucial parameter that directly shapes the system’s dynamics. Beyond a critical threshold, neuronal fluctuations increase strongly and drive the network into saturation and chaotic regimes (Fig.\ref{fig_Gen}(b-e)).  

\NI For most tasks, good performance requires staying below this threshold so that the reservoir remains in a quiescent regime, where input-induced perturbations do not develop into irregular, unpredictable fluctuations. Once the chaotic regime is entered, accuracy quickly deteriorates in the Sequence Memorization Task (SMT, Fig.\ref{fig_SMT}(b)) and in the Sequence Generation Task (SGT, Fig.\ref{fig_SGT}(f,g)).  

\NI At the same time, these two temporal tasks also require a minimal coupling strength. In the SMT, inter-neuron coupling ensures that information from the directly stimulated neurons propagates to other parts of the reservoir before it is overwritten by subsequent inputs. In the SGT, a certain degree of nonlinearity is needed to generate a dynamical landscape with sufficiently long transients into the resting-state attractor.  

\NI In the CA Prediction Task (CPT), relatively strong coupling can even be beneficial, as it enables neurons to generate binary outputs that match the predicted successor states (Fig.\ref{fig_CAT}(b)).  

\NI The Patches Classification Task (PCT) is an exception, as it does not rely on recurrent coupling at all and therefore exhibits accuracy that is independent of $w$ (Fig.\ref{fig_PCT}(a)).  

\subsubsection*{Scaling Factor $s$}

\NI The steepness of the sigmoidal neural response function in our study is controlled by the scaling factor $s$, which in other models is often interpreted as an inverse temperature. Across most of the tasks considered, accuracy exhibits a unimodal dependence on $s$, with a single peak at an optimal value (Fig.\ref{fig_SMT}(c), (Fig.\ref{fig_PCT}(c,d), Fig.\ref{fig_CAT}(c)). This indicates that a certain degree of nonlinearity is beneficial, whereas overly steep activation functions push the reservoir into harmful large-amplitude fluctuations, resembling the instability observed at large coupling strengths $w$. 

\NI In the case of the PCT, we demonstrated in detail how the scaling factor $s$ controls the deformability of decision boundaries in the reservoir computer’s input space, allowing them to align with hard class boundaries (Fig.\ref{fig_PCT}(e-h)).

\subsubsection*{Reservoir Size $N$}

\NI We found that accuracy increases monotonically with the number of reservoir neurons $N$ across all tasks in which this parameter was varied (Fig.\ref{fig_SMT}(e), Fig.\ref{fig_PCT}(b), Fig.\ref{fig_CAT}(d,e)). For large reservoirs, performance reached a perfect level of $A\approx1$ in some tasks (Fig.\ref{fig_PCT}(b), Fig.\ref{fig_CAT}(d)), whereas in others accuracy saturated at suboptimal levels (Fig.\ref{fig_SMT}(e), Fig.\ref{fig_CAT}(e)).

\subsubsection*{Type of Nonlinearity}

\NI While the standard reservoir employed the tanh activation function, we also tested alternative types in two of the tasks.

\NI In the PCT, nonlinearity is essential: tanh, sin, and Gaussian activations perform comparably well, whereas linear neurons fail completely (Fig.\ref{fig_PCT}(d)). Interestingly, Heaviside activations achieve only relatively low accuracy, suggesting that continuously deformable decision boundaries are more effective in this task, despite the presence of sharp class boundaries.

\NI In the SMT, tanh and sin activations again perform comparably (Fig.\ref{fig_SMT}(f)), but Gaussian activations perform very poorly, and Heaviside functions fail completely. In contrast, linear neurons reach perfect accuracy, indicating that old input vectors are best retrieved when reservoir states are transformed in an affine-linear way at each time step.

\subsubsection*{Division of Labor}

\NI In reservoir computing, the nonlinear recurrent network is typically regarded as the central component. The reservoir is assumed to provide the key functions of memory and data transformation, while the readout layer merely selects useful features from the reservoir’s state vector. Some of the tasks considered in this work, however, challenge this primacy of the reservoir.  

\NI An extreme case is the PCT. Although the class distribution in the two-dimensional input plane is highly complex (Fig.\ref{fig_PCT}(e)), the task relies solely on the input matrix and the nonlinearity of the activation functions. Recursive updates are not required, such that the reservoir neurons can even be left completely unconnected. In this setup, the readout layer receives only tanh-transformed linear combinations of input signals (generated by the dense input matrix) and can still separate the classes effectively.  

\NI The SGT provides another example where the main workload lies in the readout layer. To generate input-dependent predefined sequences of output states, the reservoir only needs to produce any deterministic, input-dependent state sequence, as long as subsequent states are sufficiently distinct (i.e., not linearly dependent) and do not overlap between input classes (Fig.\ref{fig_SGT}(c,f)). The readout layer can then transform such an arbitrary state sequence into the desired target sequence. This is mathematically straightforward, yet it illustrates the remarkable power of affine-linear transformations optimized via the pseudo-inverse method.  

\NI In the CPT, the workload is more balanced: both the soft nonlinearity of the activation functions and the recurrent updates are required to predict the next cellular automaton state (Fig.\ref{fig_CAT}). Nevertheless, it remains the task of the pseudo-inverse–trained readout layer to extract the local update rules of the cellular automaton from the complex reservoir state.  

\NI Taken together, our results suggest that reservoir computing does not rely on a single dominant mechanism but instead on a flexible distribution of functions across input, reservoir, and readout components. Depending on the task, the reservoir may serve as a source of memory traces, as a nonlinear feature generator, or merely as a passive conduit of transformed inputs. This perspective shifts the focus away from optimizing reservoir dynamics alone and toward understanding how different architectural elements share computational roles. In future work, it will be important to explore how these task-dependent divisions of labor extend to more complex benchmarks, and whether biologically inspired principles such as modularity, sparsity, or plasticity can be harnessed to design more efficient and interpretable reservoir systems.

\section{Additional Information}

\subsection{Author contributions}

CM conceived the study, implemented the methods, evaluated the data, and wrote the paper. AS discussed the results and acquired funding. TK and AM discussed the results and provided resources. PK conceived the study, discussed the results, acquired funding and wrote the paper.

\subsection{Funding}
This work was funded by the Deutsche Forschungsgemeinschaft (DFG, German Research Foundation): grants KR\,5148/3-1 (project number 510395418), KR\,5148/5-1 (project number 542747151), KR\,5148/10-1 (project number 563909707) and GRK\,2839 (project number 468527017) to PK, and grants SCHI\,1482/3-1 (project number 451810794) and SCHI\,1482/6-1 (project number 563909707) to AS.

\subsection{Competing interests statement}
The authors declare no competing interests.

\subsection{Data availability statement} 
The complete data and analysis programs will be made available upon reasonable request.

\subsection{Third party rights}
All material used in the paper are the intellectual property of the authors.



\newpage

\newpage

\newpage

\newpage

\newpage

\end{document}